\documentclass{article}
\usepackage{spconf,amsmath,graphicx}
\usepackage{cite}
\usepackage{amssymb,amsfonts,amsthm,bm}
\usepackage[usenames,dvipsnames]{xcolor}
\usepackage{subfigure}
\usepackage{booktabs}
\usepackage{graphicx}
\usepackage{multirow}
\usepackage{stfloats}
\usepackage{array} 
\usepackage{caption}
\usepackage{algorithm}

\usepackage{algorithmic}

\definecolor{mygreen}{rgb}{0,0.5,0}
\definecolor{darkblue}{RGB}{0,0,150}
\title{Adaptive Confidence Multi-View Hashing for Multimedia Retrieval}
\name{Jian Zhu$^{1,2}$, Yu Cui$^{2}$, Zhangmin Huang$^{2}$, Xingyu Li$^{3}$, Lei Liu$^{1,\ast}$, Lingfang Zeng$^{2,\ast}$, Li-Rong Dai$^{1}$ \thanks{$^{\ast}$Corresponding author. Lei Liu (liulei13@ustc.edu.cn) and Lingfang Zeng (zenglf@zhejianglab.com).}} 
\address{$^{1}$  University of Science and Technology of China, Hefei, China \{liulei13, lrdai\}@ustc.edu.cn\\ $^{2}$ Zhejiang Lab, Hangzhou, China \{qijian.zhu, cui.yu, zmhuang, zenglf\}@zhejianglab.com\\	$^{3}$  Lin Gang Laboratory, Shanghai, China \{lixingyu0404\}@lglab.ac.cn}
\vspace{-0.35cm}
\begin{document}
\ninept
\maketitle
\ninept
\begin{abstract}
The multi-view hash method converts heterogeneous data from multiple views into binary hash codes, which is one of the critical technologies in multimedia retrieval. However, the current methods mainly explore the complementarity among multiple views while lacking confidence in learning and fusion. Moreover, in practical application scenarios, the single-view data contains redundant noise. To conduct confidence learning and eliminate unnecessary noise, we propose a novel Adaptive Confidence Multi-View Hashing (ACMVH) method. First, a confidence network is developed to extract useful information from various single-view features and remove noise information. Furthermore, an adaptive confidence multi-view network is employed to measure the confidence of each view and then fuse multi-view features through a weighted summation. Lastly, a dilation network is designed to further enhance the feature representation of the fused features. To the best of our knowledge, we pioneer the application of confidence learning into the field of multimedia retrieval. Extensive experiments on two public datasets show that the proposed ACMVH performs better than state-of-the-art methods (maximum increase of $3.24\%$). The source code is available at https://github.com/HackerHyper/ACMVH.
\end{abstract}

\section{Introduction}
Due to the advantages of fast retrieval speed and low storage resources, hash representation learning \cite{han:64, han:63, zheng:62, zhu2024fast, welling:50, zhu2023deep, lu:18, zhu2023central} is widely used in the field of multimedia retrieval. 
Multi-view hashing utilizes the fusion of data from various views to generate a binary hash code with stronger semantic expression capabilities. How to effectively integrate multi-view data is an important research direction.

Current multi-view hashing methods suffer from the issue of untrustworthy fusion. The main reasons are detailed as follows. First, the single-view data generally contain some redundant noise features. For instance, Flexible Graph Convolutional Multi-modal Hashing (FGCMH) \cite{lu:18} is a GCN-based \cite{welling:50} multi-view hashing method. It first constructs the edges of the graph based on similarity and then the GCN aggregates features of adjacent nodes. Unfortunately, the noisy features of neighboring nodes are also introduced and aggregated to generate new features of the nodes during this procedure. Therefore, it becomes necessary to remove noise and help the multi-view hashing method achieve better performance. Second, multi-view feature fusion lacks a measure of the confidence of single-view features and the importance of measuring the confidence of each single view is underrated. To get a global representation, typical multi-view hashing methods such as Bit-aware Semantic Transformer Hashing (BSTH) \cite{tan:61} use a simple sum operation to fuse the multi-view features. However, the view-level confidence is ignored during the fusing process, which incurs a weak expressiveness of fused features. The facts above result in the problem of untrustworthy fusion.

\begin{figure}
	\centering
	\includegraphics[width=8cm]{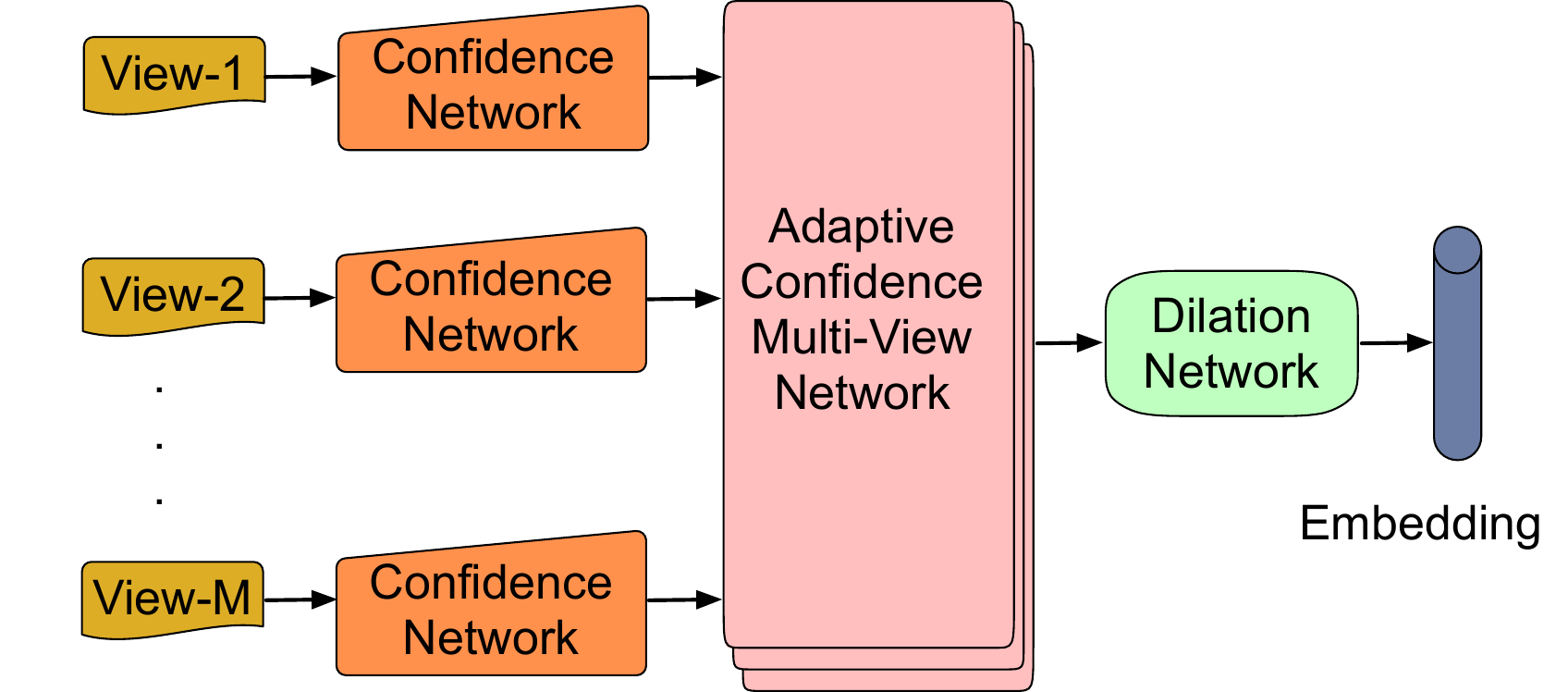}
	\caption{Adpative Confidence Multi-View Learning. Firstly, perform confidence network on individual view features to extract useful features and suppress redundant features. Secondly, automatically learn the confidence values of each single view feature and then fuse these features by a weighted summation. Finally, a dilation network is implemented on the fused feature to generate global representation.}
	\label{fig:01}
\end{figure}

To eliminate the redundant noise information and realize the confidence multi-view learning \cite{han:64, han:63, zheng:62}, we propose a novel multi-view hashing method termed \textit{Adaptive Confidence Multi-View Hashing} (ACMVH). As shown in Fig. \ref{fig:01}, adaptive confidence multi-view learning aims to learn an effective representation for the multi-view hashing task. Firstly, we utilize a confidence network for the single-view feature extraction, which can sift through useful features and suppress noise features in each single view. Then, an adaptive confidence multi-view network is used to implement confidence fusion, which can automatically learn the confidence of each view. Furthermore, based on the learned confidence of each view, we can obtain the trustworthy fusion through a weighted summation. Finally, we develop a dilation network to perform on the fused representation and enhance semantic representation further. 

We evaluate the proposed ACMVH method on MIR-Flickr25K and NUS-WIDE datasets in multi-view hash representation learning benchmarks. Our ACMVH yields an improvement of up to $3.24\%$ in mean average precision (mAP), according to benchmark results. Our main contributions are summarized as follows:
\begin{itemize}
\item To the best of our knowledge, this paper is the first to apply the confidence learning to the multi-view retrieval tasks.
\item We conduct experiments to validate the efficiency of our method and achieve SOTA results in multimedia retrieval.
\end{itemize}

\begin{figure*}
	\centering
	\includegraphics[width=15cm]{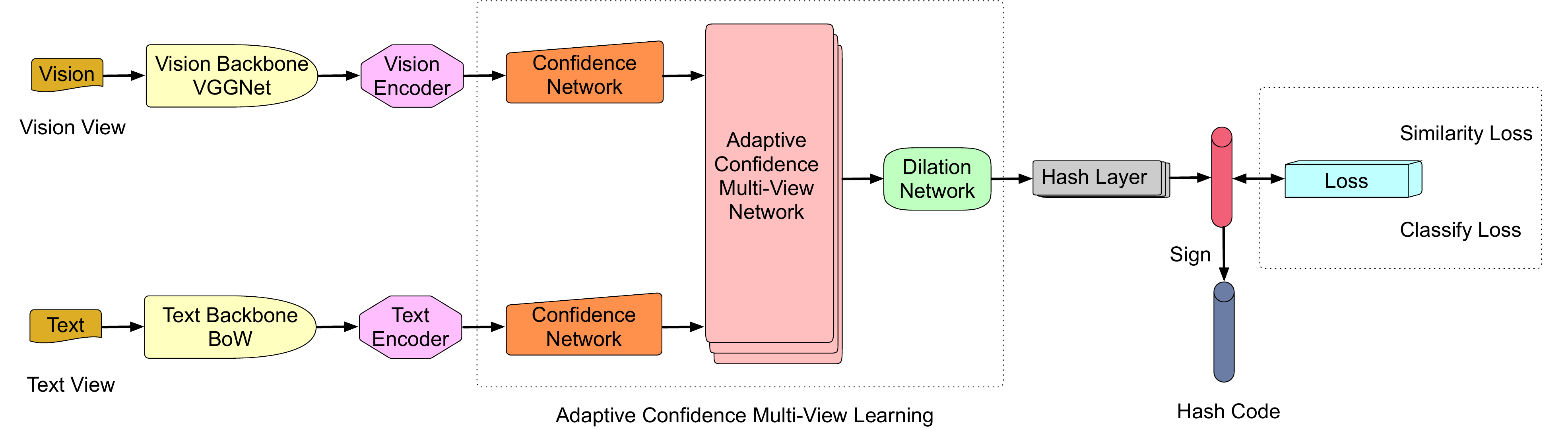}
	\caption{The flow chart of ACMVH method. The vision and text features are extracted by backbones respectively. Each single view feature needs to be mined for useful information through the confidence network. Then, view-level adaptive confidence learning is performed, and multiple view features are adaptively fused. Subsequently, the dilation network is performed on fused features to enhance the semantic representation. Finally, the hash layer outputs the binary hash codes based on the enhanced semantic representation.}
	\label{fig:02}
\end{figure*}

\section{The Proposed Methodology}
\label{section:proposed_method}
We propose adaptive confidence multi-view learning (ACMVL) to credibly fuse multi-view features and apply ACMVL to the field of multimedia retrieval. In this section, we detail the neural structure of our method and its objective.

\subsection{Deep Multi-view Hashing Network}
The deep multi-view hashing network transforms multi-view data into binary hash code. As shown in Fig. \ref{fig:02}, ACMVH consists of (1) backbones, (2) confidence networks, (3) an adaptive confidence multi-view network, (4) a dilation network, and (5) a hash layer.

\subsubsection{Backbones}
Let the training dataset be $\mathcal{X}= \left\{\left\{x_{i}\right\}_{i=1}^{N}, Y\right\}$, where $x_{i} \in \mathbb{R}^{D}$ is a multi-view instance, $N$ represents the number of samples. $Y = \left\{y_{1}, y_{2}, \ldots, y_{N}\right\}$ is a sequence set, where $y_{i}$ denotes the category information of $x_{i}$. We set $x_i = \left\{x^{1}_{i}, x^{2}_{i}, \ldots, x^{M}_{i}\right\}$ and $M$ is the number of views. Assume that
\begin{equation}
	Z^{m}_{i} =Backbone^{m}(x^{m}_{i}),
\end{equation}
$x^{m}_{i}$ represents the original data of $m$-th view. The $m$-th view data has a backbone network responsible for its respective feature $Z^{m}_{i}$. In our experiments, we utilize VGGNet \cite{simonyan:51} for vision feature extraction and Bag-of-Words model \cite{zhang:53} for text feature extraction. 

Then, we utilize a two-layer fully connected network as an encoder module. First, it can represent each view feature at a high level. Next, each view feature is normalized to the same dimension and threshold. Let
\begin{equation}
	E^{m}_{i} =Encoder^{m}(Z^{m}_{i}),
\end{equation}
where $E^{m}_{i} \in \mathbb{R}^{d}$ represents the extracted features of the sample through the encoder module in the $m$-th view and $d$ denotes the embedding dimension of the $m$-th view.

\subsubsection{Confidence Networks}
To reduce the influence of noise features, we propose a confidence network for each view to extract useful features and eliminate noise features, which improves feature confidence in each view. Let
\begin{equation}
	w^{m}_{i} =\sigma(w_{c}E^{m}_{i}+b_{c}),
\end{equation}
where $\sigma$ refers to the sigmoid activation function, $w_{c} \in \mathbb{R}^{d \times d}$ and $b_{c} \in \mathbb{R}^{d}$ are trainable parameters. The vector of weights $w^{m}_{i} \in [0, 1]^{d}$ represents a set of learned gates applied to the individual dimensions of the encoded feature $E^m_{i}$. With the learned weight vector $w^m_{i}$, the filtered features are obtained by the element-wise production between the encoded feature $E^m_{i}$ and 
the weight vector $w^m_{i}$ for each sample in each view as:
\begin{equation}
	C^{m}_{i} = w^{m}_{i} \odot E^{m}_{i}.
\end{equation}
To recap, the confidence network transforms the backbone feature $Z^m_{i}$ into a new representation $C^{m}_{i}$.

\subsubsection{Adaptive Confidence Multi-View Network}
The importance of individual views varies in multimedia retrieval tasks. We learn the confidence value of each view as:
\begin{equation}
	A_{i}  = \sum_{m =1}^{M} (p^{m} * C^{m}_{i}),
\end{equation}
where $p^{m}$ represents the confidence weight of $m$-th view. In fact, $p^{m}$ is also a part of the neural network parameters, therefore, the optimal $p^{m}$ can be obtained through training. Further, $A_{i}$ is the result of the confidence fusion of multiple view features.

\subsubsection{Dilation Network}
Lastly, a dilation network structure is developed for the semantic enhancement of fused features. This module first increases the dimension of the fused features and then reduces them to the original dimension. More precisely, the specific structure consists of two layers: $U_{i}$ and $G_{i}$. First, $U_{i}$ is defined as follows:
\begin{equation}
U_{i} = ReLU(w_{u1}A_{i} + b_{u1}),
\end{equation}
where $ReLU$ refers to the ReLU activation function, $w_{u1} \in \mathbb{R}^{d \times 4d}$ and $b_{u1} \in \mathbb{R}^{4d}$ are deep network parameters.
Then, we define $G_{i}$ by
\begin{equation}
G_{i} =w_{u2}U_{i} + b_{u2} + A_{i}
\end{equation}
$w_{u2} \in \mathbb{R}^{4d \times d}$ and $b_{u2} \in \mathbb{R}^{d}$ are trainable parameters. Notice that, $G_{i}$ is the final global representation of multi-view features.

\subsubsection{Hash Layer}
A linear layer with a tanh activation is hired as the hash layer, which can be represented as:
\begin{equation}
h_i = \tanh(w_{h}G_i+b_{h}), h_i \in \mathbb{R}^{1 \times k},
\end{equation}

\begin{equation}
b_i=\operatorname{sign}\left({h_i}\right), b_i \in\{-1,1\}^{1 \times k},
\end{equation}
where $sign$ is the signum function, $w_{h} \in \mathbb{R}^{d \times k}$ and $b_{h} \in \mathbb{R}^{k}$ are network parameters. $k$ indicates that the hash layer generates $k$-bit hash code.

\subsection{Loss Functions}
The loss function shown below is used to learn the hash codes while taking the similarity metric between samples into account:
\begin{equation}
{L}_{sim}=\left\|\cos \left({h}_i, {h}_j\right)-{\phi}_{i j}\right\|_2^2,
\label{eq:01}
\end{equation}
where $\phi$ is the affinity matrix, which can model the relation between relevant samples. $\phi_{i j}$ is calculated as follows:
\begin{equation}
{\phi}_{i j}=\frac{2}{1+e^{-y_i y_j^T}}-1.
\end{equation}
Notice that, the category information is not completely utilized even if pairwise category information is used to train the hash function in Eq.\eqref{eq:01}. 
We believe that the learned binary codes should be suitable for categorization. To describe the connection between the learned binary codes and the category information, we utilize a simple linear classifier. 
The classifier loss function can be formulated as:
\begin{equation}
{L}_{clf}=\left\|{y'}_i-{y}_{i}\right\|_2^2,
\end{equation}
where 
\begin{equation}
{y'}_{i}=Linear(h_i),
\end{equation}
is the predicted value of the linear classifier and the squared L2 norm is used as the loss for classification.

We can derive the total loss function as
\begin{equation}
	L_{total}= L_{sim}+\mu L_{clf},
	\label{eq:loss}
\end{equation}
where $\mu$ is the hyper-parameter obtained through grid search in our work. 

\begin{table*}
	\centering
        \caption{General statistics of two datasets. The dataset size, number of categories, and feature dimensions are included.}
	\begin{tabular}{llllllll}
		\toprule[1pt]
		Dataset   & Training Size & Retrieval Size & Query Size & Categories&Visual Embedding & Textual Embedding \\ \midrule[0.8pt]
		MIR-Flickr25K & 5000  & 17772 & 2243    & 24&4096-D & 1386-D \\
  NUS-WIDE & 21000  & 193749 & 2085    & 21&4096-D &1000-D\\
		
		\bottomrule[1pt]
	\end{tabular}
	
	\label{Tab:01}
\end{table*}

\section{Experiments}
\subsection{Evaluation Datasets and Metrics}
We evaluate the proposed ACMVH method on multimedia retrieval tasks in experiments. Two public datasets are selected: MIR-Flickr25K \cite{huiskes:20} and NUS-WIDE \cite{chua:22}. These datasets are widely used for evaluating multimedia retrieval performance. We use the mean Average Precision (mAP) as the evaluation metric. The details of two datasets used in experiments are summarized in Table \ref{Tab:01}.
\subsection{Baseline}
To evaluate the retrieval metric, we compare the proposed ACMVH method with six multi-view hashing methods (e.g., Flexible Discrete Multi-view Hashing (FDMH) \cite{liu:23}, Flexible Online Multi-modal Hashing (FOMH) \cite{lu:24}, Deep Collaborative Multi-View Hashing (DCMVH) \cite{zhu:17}, Supervised Adaptive Partial Multi-view Hashing (SAPMH) \cite{zheng:25}, Flexible Graph Convolutional Multi-modal Hashing (FGCMH) \cite{lu:18}, and Bit-aware Semantic Transformer Hashing (BSTH) \cite{tan:61}).

\begin{table*}
	\setlength{\tabcolsep}{2pt}
	\centering
	\caption{The comparable mAP results on MIR-Flickr25K and NUS-WIDE. The best results are bolded, and the second-best results are underlined. The * indicates that the results of our method on this dataset are statistically significant.}
    \begin{tabular}{llllllllllllll}
		\toprule[1pt]
		\multicolumn{1}{c}{\multirow{2}{*}{Method}} & \multicolumn{1}{c}{\multirow{2}{*}{Ref.}} & \multicolumn{4}{c}{MIR-Flickr25K*}    & \multicolumn{4}{c}{NUS-WIDE*}            \\  \cmidrule(r){3-6}  \cmidrule(r){7-10}  
		\multicolumn{1}{c}{}                         & \multicolumn{1}{c}{}                      & 16 bits & 32 bits & 64 bits & 128 bits & 16 bits & 32 bits & 64 bits & 128 bits  \\ \midrule[0.8pt]
		FOMH                                         & MM19                                      & 0.7557 & 0.7632 & 0.7564 & 0.7705  & 0.6329 & 0.6456 & 0.6678 & 0.6791    \\
		FDMH                                         & NPL20                                     & 0.7802 & 0.7963 & 0.8094 & 0.8181  & 0.6575 & 0.6665 & 0.6712 & 0.6823    \\
		DCMVH                                        & TIP20                                     & 0.8097 & 0.8279 & 0.8354 & 0.8467  & 0.6509 & 0.6625 & 0.6905 & 0.7023    \\
		SAPMH                                        & TMM21                                      & 0.7657 & 0.8098 & 0.8188 & 0.8191  & 0.6503 & 0.6703 & 0.6898 & 0.6901   \\
		FGCMH                                        & MM21                                      & \underline{0.8173} & \underline{0.8358} & 0.8377 & \underline{0.8606}  & 0.6677 & 0.6874 & 0.6936 & 0.7011   \\
  BSTH & SIGIR22  &0.8145&0.8340&\underline{0.8482}&0.8571                                     & \underline{0.6990} & \underline{0.7340} & \underline{0.7505} & \underline{0.7704}   \\\midrule[0.8pt]
		ACMVH                                         & Proposed                                         &\textbf{0.8424} & \textbf{0.8573} & \textbf{0.8692} & \textbf{0.8740} & \textbf{0.7314} & \textbf{0.7562} & \textbf{0.7719} & \textbf{0.7762}   \\
		\bottomrule[1pt]
	\end{tabular}

	\label{Tab:02}
\end{table*}

\subsection{Analysis of Experimental Results}

\begin{table*}[!t]
	\setlength{\tabcolsep}{2pt}
	\centering
        \caption{Ablation Experiments On Two Datasets. Effects of Adaptive Confidence Multi-View Hash Architecture.}
	\begin{tabular}{lllllllllllll}
		\toprule[1pt]
		\multicolumn{1}{c}{\multirow{2}{*}{Methods}} & \multicolumn{4}{c}{MIR-Flickr25K}   & \multicolumn{4}{c}{NUS-WIDE}   \\   \cmidrule(r){2-5}  \cmidrule(r){6-9}  
		\multicolumn{1}{c}{} & 16 bits & 32 bits & 64 bits & 128 bits & 16 bits & 32 bits & 64 bits & 128 bits \\  \midrule[0.8pt] 
		
		ACMVH-text    & 0.6921 &  0.7047 &  0.7201 &  0.7252  & 0.5876 & 0.6069 & 0.6308 & 0.6474 \\
		ACMVH-vision    & 0.8076 &  0.8213 &  0.8356 &  0.8464  & 0.6677 & 0.7041 &0.7252 & 0.7413     \\
        ACMVH-concat    & 0.8108 &  0.8235 &  0.8442 &  0.8542  & 0.6924 & 0.7235 & 0.7576 & 0.7621  \\ 
            ACMVH-adaptive    & 0.8262 &  0.8390 &  0.8548 &  0.8662  & 0.7150 & 0.7455 & 0.7613 & 0.7721\\
		ACMVH-confidence    & 0.8109 &  0.8341 &  0.8509 &  0.8587  & 0.7007 & 0.7339 & 0.7599 & 0.7710 \\
          ACMVH-dilation    & 0.8317 &  0.8394 &  0.8583 &  0.8670  & 0.7222 & 0.7512 & 0.7653 & 0.7735 \\ \midrule[0.8pt]
		ACMVH     &\textbf{0.8424} & \textbf{0.8573} & \textbf{0.8692} & \textbf{0.8740} & \textbf{0.7314} & \textbf{0.7562} & \textbf{0.7719} & \textbf{0.7762}  \\
		\bottomrule[1pt]
	\end{tabular}
	
	\label{Tab:03}
\end{table*}
\begin{figure}
  \centering
  \subfigure{\includegraphics[scale=0.21]{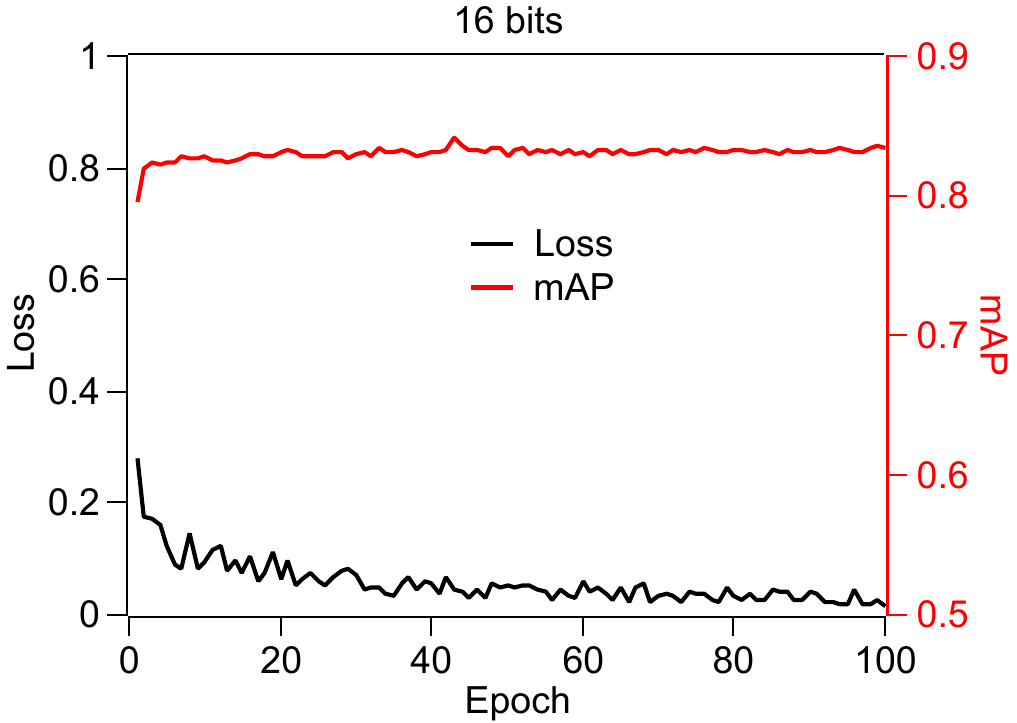}}
  \subfigure{\includegraphics[scale=0.21]{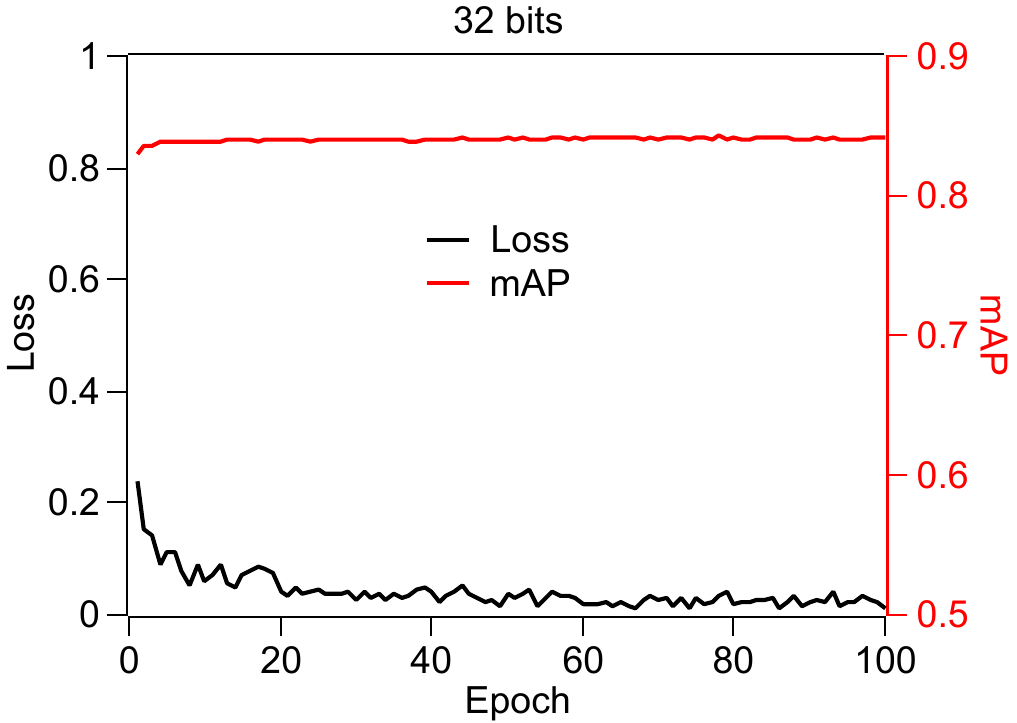}}
  \subfigure{\includegraphics[scale=0.21]{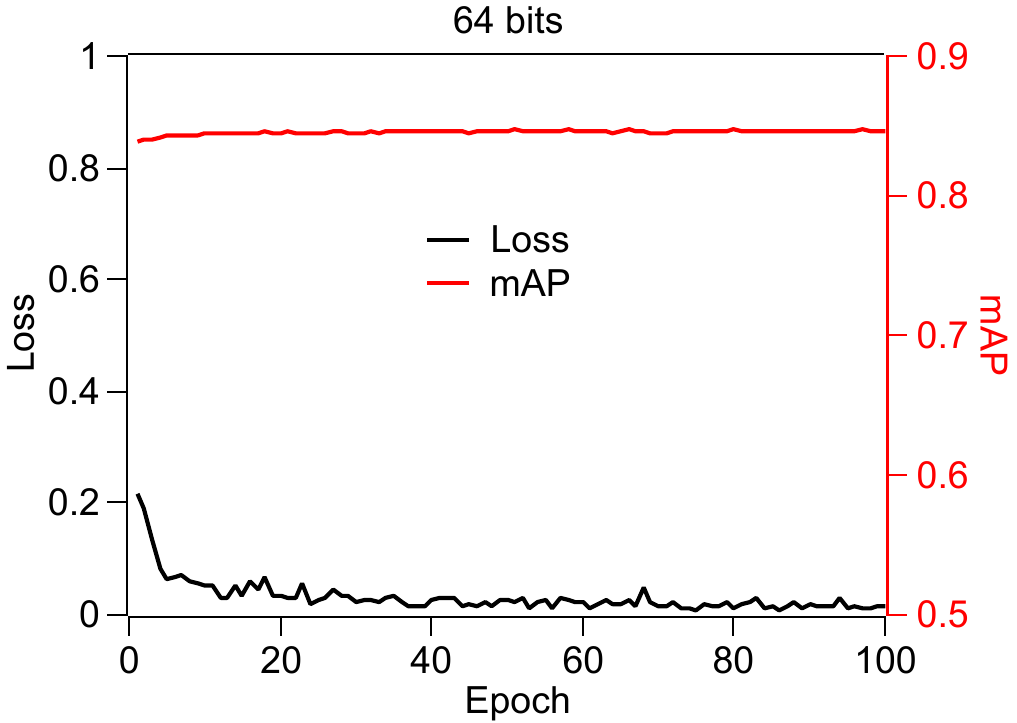}}
  \subfigure{\includegraphics[scale=0.21]{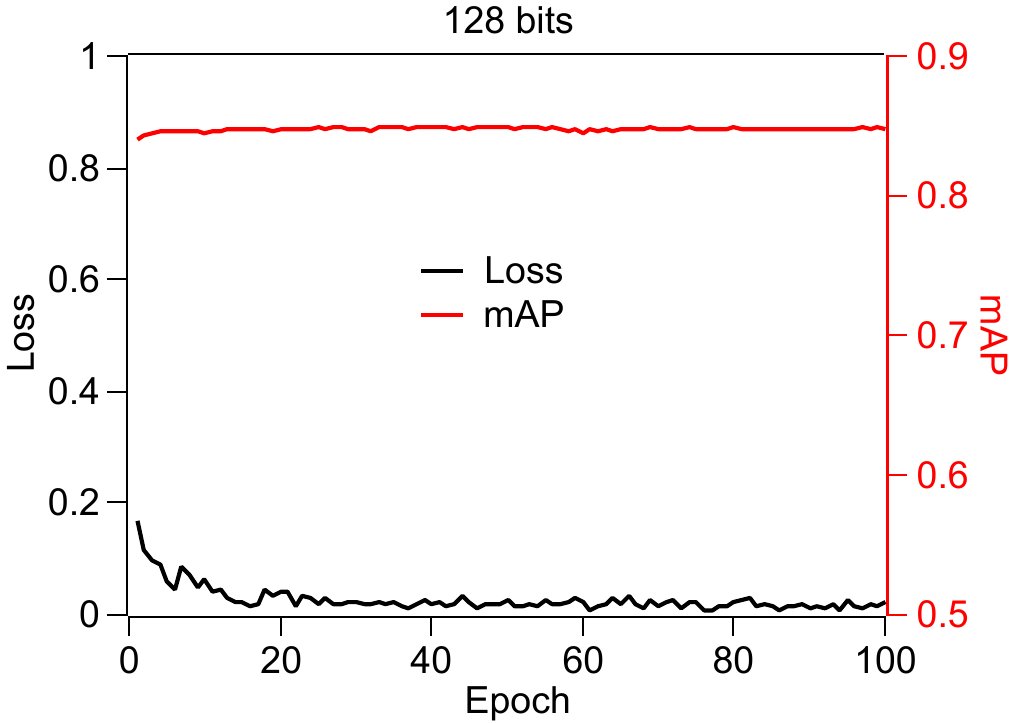}}
  \caption{The training loss and test mAP curves on MIR-Flickr25K dataset.}
  \label{fig:03}
\end{figure}
The experimental comparisons of all methods are conducted according to the unified conditions of the train set, the retrieval set, and the query set in Table \ref{Tab:01}. All multi-view hashing methods use the same backbone networks to extract visual and textual features.

The mAP result is shown in Table \ref{Tab:02}. The results show that the proposed ACMVH  is overall better than all the compared multi-view hashing methods by a large margin. For example, compared with the current state-of-the-art multi-view hashing method Bit-aware Semantic Transformer Hashing (BSTH) \cite{tan:61}, the average mAP score of our method has increased by $2.12\%$, and $2.05\%$ on MIR-Flickr25K and NUS-WIDE, respectively. The main reasons for these superior results come from three aspects: 
\begin{itemize}
\item The confidence network can extract useful features of a single view effectively and suppress noise features. 
\item Adaptive confidence multi-view network could credibly fuse the multi-view features into a global representation. 
\item Dilation network enhances the semantic representation of fused multiple view embedding. 
\end{itemize}
Adaptive confidence multi-view learning promotes the discriminative capability of hash codes.

\subsection{Ablation Studies}
To evaluate our method component by component, we perform an ablation of the proposed ACMVH with different experiment settings and report the performance. The experiment settings are as follows:
\begin{itemize}
	\item \emph{ACMVH-text}: Only the text feature is used for retrieval.
	\item \emph{ACMVH-vision}: Only the vision feature is used for retrieval.
	\item \emph{ACMVH-concat}: Vison and text features are fused with concatenation without adaptive confidence multi-view learning.
        \item \emph{ACMVH-adaptive}: The adaptive confidence multi-view network is removed.
	\item \emph{ACMVH-confidence}: The confidence network is removed. 
     \item \emph{ACMVH-dilation}: The dilation network is removed. 
	\item \emph{ACMVH}: Our full method.
\end{itemize}
The comparison results are presented in Table \ref{Tab:03}. ACMVH-vision performs better than ACMVH-text in all tasks by a large margin indicating the vision features contain more useful information than text. By comparing ACMVH-concat with ACMVH-vision, we performed a basic concatenation of visual and text features to achieve a slight performance improvement. ACMVH is the full use of our adaptive confidence multi-view learning, which greatly improves the performance of mAP compared to ACMVH-concat. Based on the performance of ACMVH-adaptive, ACMVH-confidence, and ACMVH-dilation, it is evident that the confidence network holds the highest significance, followed by the adaptive confidence multi-view network, and finally, the dilation network ranks last.

\subsection{Convergence Analysis}
To verify the generalization performance and convergence of ACMVH, we conduct some experiments. We run hash benchmarks with varying code lengths on the MIR-Flickr25K dataset. Fig. \ref{fig:03} shows training loss and test mAP. As the training goes on, the loss steadily decreases. The loss is steady after $60$ epochs, proving that the local minimum reaches. The mAP for the test metric rapidly rises when the experiment begins. After $40$ epochs, the test mAP stays stable. Further training does not result in a deterioration of the test MAP, indicating good generalization capability. We observe similar results for different datasets

\section{Conclusion and Future work}
To enhance the feature representation, adaptive confidence multi-view learning (ACMVL) is developed.
Under multiple experiment settings, it delivers up to $3.24\%$ performance gain over the current state-of-the-art methods. However, we notice some issues, for instance,
the performance gain is not quite significant as the length of the hash
code increases. We will work on these issues to further improve the
proposed method.

\section{Acknowledgment}
This work is supported in part by the Zhejiang provincial ``Ten Thousand Talents Program'' (2021R52007), the National Key R\&D Program of China (2022YFB4500405), the Science, Technology Innovation 2030-Major Project (2021ZD0114300), Sponsored by Zhejiang Lab Open Research Project (NO.K2022QA0AB04) , Supported by ``the Fundamental Research Funds for the Central Universities''  and supported by the National Natural Science Foundation of China (62306291).
\vspace{-9pt}

\ninept 

\bibliographystyle{IEEEtran}

\bibliography{IEEEabrv,myabrv_new,refs}
\end{document}